\pdfoutput=1

\documentclass[11pt]{article}

\usepackage[final]{acl}

\usepackage{times}
\usepackage{latexsym}

\usepackage[T1]{fontenc}

\usepackage[utf8]{inputenc}

\usepackage{microtype}

\usepackage{inconsolata}

\usepackage{xspace}
\usepackage{graphicx}
\usepackage{caption}
\usepackage{multirow}
\usepackage{multicol}
\usepackage{makecell}
\usepackage{booktabs}
\usepackage{tabularx}
\usepackage{adjustbox}
\usepackage{pifont}
\usepackage{soul}
\usepackage{enumitem}
\usepackage{amsmath}
\usepackage{amssymb}
\usepackage{comment}
\usepackage{enumitem}
\usepackage{cleveref}
\usepackage{arydshln}

\crefformat{section}{\S#2#1#3}
\crefformat{subsection}{\S#2#1#3}
\crefformat{subsubsection}{\S#2#1#3}
\crefrangeformat{section}{\S\S#3#1#4 to~#5#2#6}
\crefmultiformat{section}{\S\S#2#1#3}{ and~#2#1#3}{, #2#1#3}{ and~#2#1#3}

\usepackage{graphicx,dblfloatfix}

\title{\textsc{Pearl}: A Review-Driven Persona-Knowledge Grounded \\ Conversational Recommendation Dataset}

\author{
        Minjin Kim\textsuperscript{\rm 1}\thanks{Equal contribution}~~~
        Minju Kim\textsuperscript{\rm 1}$^\ast$~~~
        Hana Kim\textsuperscript{\rm 1}~~~
        Beong-woo Kwak\textsuperscript{\rm 1}~~~ \\
        \textbf{SeongKu Kang}\textsuperscript{\rm 3}~~~
        \textbf{Youngjae Yu}\textsuperscript{\rm 1}~~~
        \textbf{Jinyoung Yeo}\textsuperscript{\rm 1}~~~
        \textbf{Dongha Lee}\textsuperscript{\rm 1}\thanks{Corresponding author}\\
        Yonsei University, Korea\textsuperscript{\rm 1}\\
        University of Illinois at Urbana-Champaign, USA\textsuperscript{\rm 3}\\
        \texttt{\{minjin.kim,minnju,yjy,jinyeo,donalee\}@yonsei.ac.kr}~~~\texttt{seongku@illinois.edu}
        }

\begin{document}
\maketitle
\newcommand{\todoc}[2]{{\textcolor{#1}{#2}}}
\newcommand{\todoblue}[1]{\todoc{blue}{#1}}

\newcommand{\todocc}[2]{{\textcolor{#1}{[[#2]]}}}
\newcommand{\todored}[1]{\todocc{red}{[[#1]]}}
\newcommand{\hist}[1]{\todored{hist: #1}}

\newcommand{\todocccc}[2]{{\textcolor{#1}{[[#2]]}}}
\newcommand{\todogreen}[1]{\todocccc{green}{[[#1]]}}
\newcommand{\minju}[1]{\textcolor{brown}{#1}}

\newcommand{\se}{{\it SE}}%
\newcommand{\eg}{{\it e.g.}}%
\newcommand{\ie}{{\it i.e.}}%
\newcommand{\etal}{{\it et al.}}%
\newcommand{\etc}{{\it etc}}%
\newcommand{\ourframework}{\textsc{Clam}\xspace}
\newcommand{\ourdata}{\textsc{Pearl}\xspace}

\newcommand{\argmin}{\operatornamewithlimits{argmin}}
\newcommand{\argmax}{\operatornamewithlimits{argmax}}
\newcommand{\tabitem}{~~\llap{\textbullet}~~}
\definecolor{yellow-green}{rgb}{0.3, 0.5, 0.0}

\definecolor{lightergray}{RGB}{230,230,230}
\definecolor{DarkGreen}{RGB}{30,130,30}
\newcommand{\cmark}{\textcolor{DarkGreen}{\ding{51}}}
\newcommand{\xmark}{\textcolor{red}{\ding{55}}}%


\newcommand{\mcal}[1]{{\cal{#1}}}
\newcommand{\calA}{\mbox{${\cal A}$}}
\newcommand{\calB}{\mbox{${\cal B}$}}
\newcommand{\calC}{\mbox{${\cal C}$}}
\newcommand{\calD}{\mbox{${\cal D}$}}
\newcommand{\calE}{\mbox{${\cal E}$}}
\newcommand{\calF}{\mbox{${\cal F}$}}
\newcommand{\calG}{\mbox{${\cal G}$}}
\newcommand{\calH}{\mbox{${\cal H}$}}
\newcommand{\calI}{\mbox{${\cal I}$}}
\newcommand{\calJ}{\mbox{${\cal J}$}}
\newcommand{\calK}{\mbox{${\cal K}$}}
\newcommand{\calL}{\mbox{${\cal L}$}}
\newcommand{\calM}{\mbox{${\cal M}$}}
\newcommand{\calN}{\mbox{${\cal N}$}}
\newcommand{\calO}{\mbox{${\cal O}$}}
\newcommand{\calP}{\mbox{${\cal P}$}}
\newcommand{\calQ}{\mbox{${\cal Q}$}}
\newcommand{\calR}{\mbox{${\cal R}$}}
\newcommand{\calS}{\mbox{${\cal S}$}}
\newcommand{\calT}{\mbox{${\cal T}$}}
\newcommand{\calU}{\mbox{${\cal U}$}}
\newcommand{\calV}{\mbox{${\cal V}$}}
\newcommand{\calW}{\mbox{${\cal W}$}}
\newcommand{\calX}{\mbox{${\cal X}$}}
\newcommand{\calY}{\mbox{${\cal Y}$}}
\newcommand{\calZ}{\mbox{${\cal Z}$}}

\definecolor{lightblue}{RGB}{224,236,247}
\definecolor{deepblue}{RGB}{9,46,107}

\newcommand{\subscript}[2]{$#1 _ #2$}

\begin{abstract}

Conversational recommender systems are an emerging area that has garnered increasing interest in the community, especially with the advancements in large language models (LLMs) that enable sophisticated handling of conversational input. Despite the progress, the field still has many aspects left to explore. The currently available public datasets for conversational recommendation lack specific user preferences and explanations for recommendations, hindering high-quality recommendations. To address such challenges, we present a novel conversational recommendation dataset named \ourdata{}, synthesized with persona- and knowledge-augmented LLM simulators.
We obtain detailed persona and knowledge from real-world reviews and construct a large-scale dataset with over 57k dialogues. Our experimental results demonstrate that \ourdata{} contains more specific user preferences, show expertise in the target domain, and provides recommendations more relevant to the dialogue context than those in prior datasets. Furthermore, we demonstrate the utility of \ourdata{} by showing that our downstream models outperform baselines in both human and automatic evaluations. We release our dataset\footnote{\url{https://huggingface.co/datasets/DLI-Lab/pearl}} and code.\footnote{\url{https://github.com/kkmjkim/PEARL}}


\end{abstract}
\begin{figure}[t!]
    \centering
    \includegraphics[width=\linewidth]{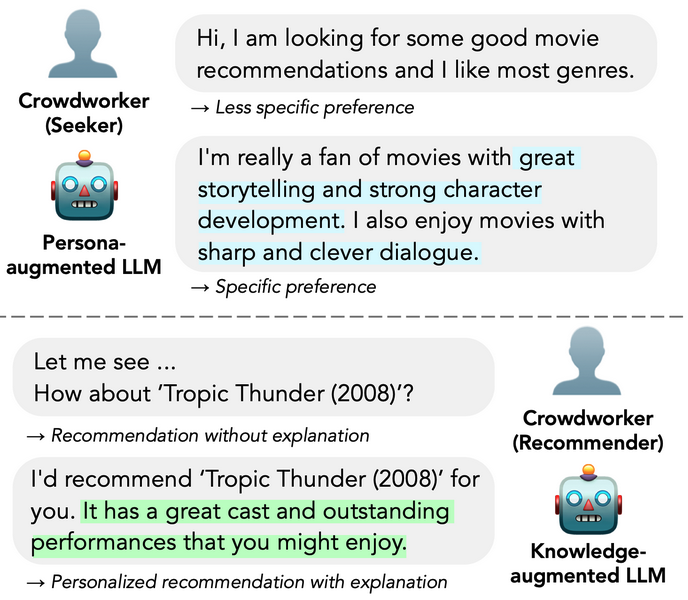}
    \caption{An example comparing utterances of crowdworkers and our persona-knowledge augmented LLM simulators.}
    \label{fig:intro}
\end{figure}

\section{Introduction}
Recently, conversational recommender system (CRS) has become an emerging research topic, which aims to elicit user preferences and offer personalized recommendations by engaging in interactive conversations. Towards this goal, an increasing emphasis has been placed on constructing high-quality dataset \citep{li2018towards,liu2020towards,zhou2020towards,hayati2020inspired}.
Existing conversational recommendation datasets are mainly collected via crowdsourcing, which is to gather interactions between two crowdworkers where one plays a role as a user (\ie, recommendation seeker) and the other pretends as a recommender.

However, existing dialogues have several limitations that impede downstream CRS models from delivering satisfactory user experiences.
First, user preferences expressed in the existing CRS datasets are often less specific. An example includes dialogues with statements like, \textit{``I like most genres''} as shown in Figure \ref{fig:intro}. Such dialogues lead downstream models to offer recommendations that are generic and less personalized 
\citep{he2023large,zhou2020improving}.
This happens as crowdworkers, playing the role of users, often lack specific preferences during their tasks, unlike most real-world users who seek personalized recommendations. 
Second, conversational recommendation dialogues often provide suboptimal recommendations and insufficient explanations alongside recommendations, due to the limited knowledge of crowdworkers \citep{guo2023towards}. A specific example is dialogues containing utterances like \textit{``Let me see... How about Tropic Thunder?''} as shown in Figure \ref{fig:intro}. Such responses can lead to CRS models generating less accurate and relevant suggestions. Additionally, the absence of explanations alongside recommendations can be a crucial hurdle, preventing users from grasping the items and the rationale behind the recommendations.

\begin{table*}[!t]
    \centering
    \resizebox{0.95\linewidth}{!}{
    \begin{tabular}{lccccc}
        \toprule
        \textbf{} & \makecell{\textbf{\ourdata}\\(this work)}& \makecell{ \textbf{ReDial} \\\citep{li2018towards} } & 
        \makecell{ \textbf{INSPIRED}\\\citep{hayati2020inspired} } &  \makecell{ \textbf{TG-ReDial} \\\citep{zhou2020towards} } & \makecell{ \textbf{DuRecDial 2.0}\\\citep{liu2021durecdial} } 
        \\
         \cmidrule(lr){1-1}  \cmidrule(lr){2-2}  \cmidrule(lr){3-3}  \cmidrule(lr){4-4}  \cmidrule(lr){5-5} \cmidrule(lr){6-6}
          
          Collection method   & Synthesized & Crowdsourced & Crowdsourced & Human-Machine & Crowdsourced  \\ 
          Real-world persona  & \cmark & \xmark & \xmark & \cmark & \xmark  \\ 
          Explained recommendation   & \cmark & \xmark & \xmark & \xmark & \xmark  \\ 
          \hdashline
          Number of dialogues       & $\mathbf{57,277}$  & $10,006$   & $1,001$   & $10,000$   & $16,482$ \\ 
          Number of users           & $\mathbf{4,680}$  & $956$   & $1,594$   & $1,482$   & $2,714$ \\ 
          Number of utterances      & $\mathbf{548,061}$ & $182,150$   & $35,811$   & $129,392$   & $255,346$ \\ 
         \bottomrule
    \end{tabular}
    }
    \caption{
    {A comparison of our synthesized dataset to notable conversational recommendation datasets. }
    }
    \label{tab:dataset_comparison}
\end{table*}

In this paper, we present a conversational recommendation dataset named \ourdata{} (Persona and knowledgE Augmented Recommendation diaLogues) that addresses the limitations of existing datasets. We transform reviews into persona and item knowledge and incorporate large language model (LLM)-based simulators augmented with persona and knowledge.
Each simulator is designed to enhance the preference specificity and informativeness of the collected data, respectively. 
The \textit{user simulator} plays the role of a recommendation seeker and is equipped with a persona, which is a set of sentences describing features that the user likes and dislikes. Each persona can help the simulator express a distinct user preference of a single real-world user as it is constructed based on the item reviews written by the same user. By simulating users with distinct preferences and specific needs, we can generate dialogues with consistent and clear preferences. 
The \textit{recommender simulator} is designed to emulate a recommender with domain knowledge. This addresses the lack of proper recommendations and explanations in existing datasets. Specifically, we incorporate item reviews which can not only provide basic information about items but also reveal soft attributes of items that can be only described through experience (\eg, \textit{``feel-good movies''} or \textit{``upbeat music''}) which could be crucial information in the users' decision-making process.

Our collected dataset includes over 57.2k dialogues simulating over 4k users and covering more than 9k items. 
Our human evaluation results show that our synthesized dataset is preferred by human raters compared to other crowdsourced dialogues. Moreover, our dataset covers a broader spectrum of user needs as we utilize a large set of authentic reviews written by diverse users.

We also conduct extensive experiments assessing the utility of \ourdata{} through human evaluation and automatic evaluation. The results demonstrate that CRS models trained on \ourdata{} show competitive or better performances in recommendation and response generation tasks compared to models trained on human-annotated datasets. Additionally, human judges consistently favor responses from models trained on \ourdata{} over those from models trained on crowdsourced datasets, across all metrics. Furthermore, our experimental results empirically validate that CRS models trained on \ourdata{} generalize better to unseen dialogues than the models trained on existing datasets.

\section{Related Work}
\subsection{Conversational Recommendation}
Conversational recommendation is an emerging task where the main goal is to provide high-quality recommendations to users through natural language conversations. Compared to traditional recommendation tasks, conversational recommendation is a more challenging task as it requires the model to not only recommend appropriate items (\ie, recommendation) but also generate engaging and helpful responses (\ie, response generation).

To facilitate the study of conversational recommendation, several datasets have been proposed by previous work. \citet{li2018towards} and \citet{hayati2020inspired} combine the elements of social chitchat and recommendation dialogues. \citet{zhou2020towards} collect human-machine conversation data guided by pre-defined topics. While most existing work has collected conversational recommendation data through crowdsourcing, those datasets are often not scalable and can suffer from less diverse user preferences and uninformative recommendations.

To handle this limitation, \citet{he2023large} scrape single-turn recommendation dialogues from Reddit. However, their practical usability has been limited due to the low quality of scraped dialogues and lack of active interactions. Similar to our work, \citet{lu2023august} generate recommendation dialogues by converting user-item interactions into dialogues with a data-to-text generation model. Nonetheless, as the data-to-text model is trained on existing recommendation dialogues, the generated dialogues still inherit the previous limitations of the crowdsourced dataset. In contrast, we tackle the limitation by fully utilizing LLMs for collecting recommendation dialogues and leveraging user reviews to enhance the diversity and informativeness of collected dialogues.


\subsection{Synthetic Data Generation}
LLMs have been increasingly used to synthesize dialogue datasets \citep{kim2022botstalk,lu2023august,chen-etal-2023-places,kim-etal-2023-soda,chae2023dialogue}.
\citet{kim2022botstalk} build Blended Skill BotsTalk using multiple agents grounded in target skills. \citet{chen-etal-2023-places} generate dyadic and multi-party conversations with topic words and show they have comparable quality to human-authored conversations. There are recent approaches that utilize external sources for generating high-quality dialogues with LLMs \citep{li-etal-2022-controllable,zhou-etal-2022-reflect,kulhanek-etal-2021-augpt}. \citet{kim-etal-2023-soda} sought to distill conversations from InstructGPT 175B using a commonsense knowledge graph. Compared to existing works, we are the first to utilize dynamic input sources for generating informative responses on a significantly large-scale, which allows us to encompass an exceptionally broad information of user preferences and item information.



\begin{figure*}[t!]
    \centering
    \includegraphics[width=\linewidth]{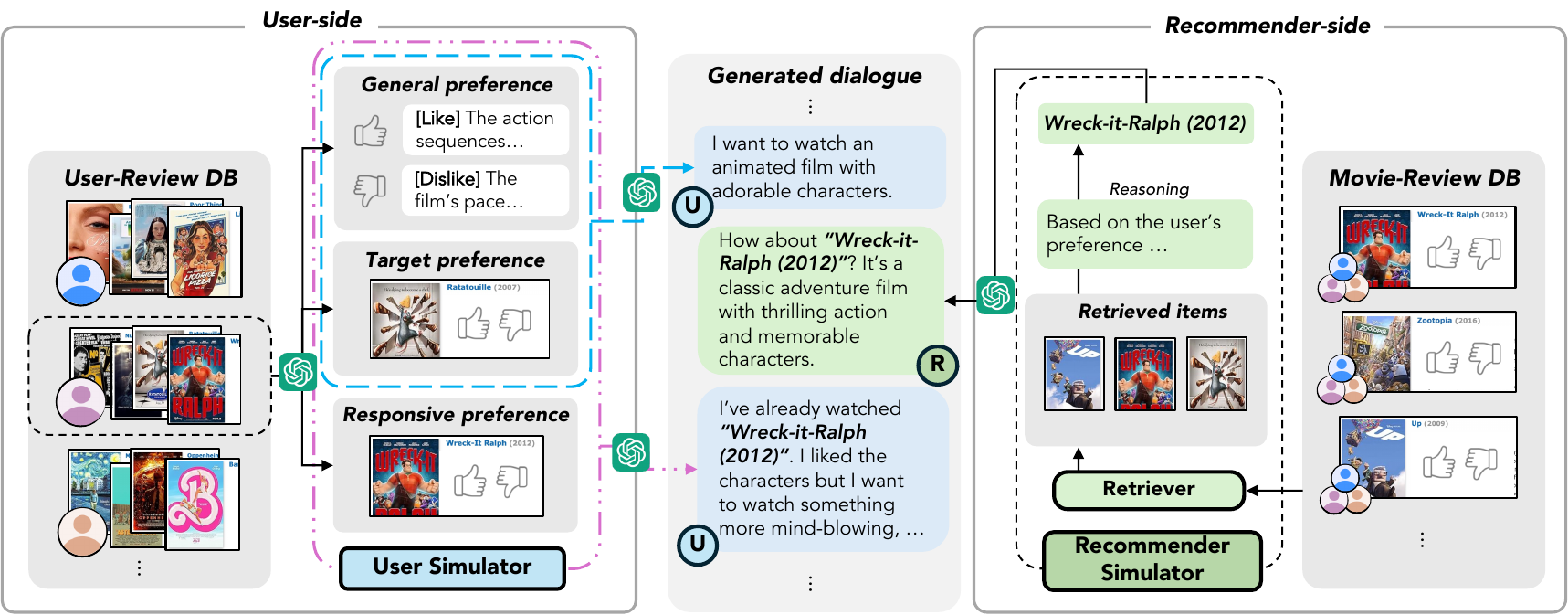}
    \caption{The overview of \ourdata{} construction method. We synthesize recommendation dialogues with review-driven persona-knowledge grounded simulators. Specifically, our user simulator is equipped with persona and our recommender simulator is augmented by knowledge derived from reviews.}
    \label{fig:method}
\end{figure*}
\section{\ourdata{} Construction}~\label{method:framework}
We construct \ourdata{}, a large-scale recommendation dataset covering diverse preferences and detailed item explanations through five steps: (1) grouping real-world reviews with two axes, which are user and item, (\cref{method:construct_db}), (2) equipping a user simulator with preferences of a real-world user using reviews (\cref{method:user_simulator}), (3) infusing item knowledge extracted from reviews to a recommender simulator (\cref{method:rec_simulator}), (4) inferring the simulators to derive a dialogue (\cref{method:gen_dialogues}), and (5) filtering dialogues (\cref{method:filtering}). While our dataset construction process is domain-independent, we validate it in the movie domain due to its extensive prior attention~\citep{li2018towards,hayati2020inspired,liu2020towards,zhou2020towards}, making it easier to evaluate. 
The overview of \ourdata{} construction process is shown in Figure \ref{fig:method} and an example of \ourdata{} is in Table \ref{tab:example}.

\subsection{Constructing User-Review and Item-Review Databases}~\label{method:construct_db}
To grant realistic preferences to the user simulator and item knowledge to the recommender simulator, we collect authentic reviews by scraping movie and review data from IMDB website.\footnote{https://www.imdb.com/}

For the user simulator, we construct a user-review database where a set of movie title, review text, and rating is grouped by the user who wrote the review. To clarify the preference and remove any noise from the raw review text, we transform the raw text into a high-level review text that focuses on the features the user likes and dislikes by using an LLM.
By utilizing a high-level review text instead of the raw text, we can help our user simulator ground on refined text without noise and also filter out personal information in the raw reviews.
The prompt is shown in Table \ref{tab:user_review_prompt}.

For the recommender simulator which should provide proper recommendations based on rich knowledge about the items, we construct an item-review database where a set of genre, director, cast, and review text is grouped per movie title. By utilizing reviews, we can obtain information about items that cannot be gathered by only utilizing metadata of items. Here, we select up to three most voted reviews on IMDB for efficiency, instead of using all collected reviews of the item. Then, we transform them into a high-level review text, similar to the user-review database construction process.
The prompt is shown in Table \ref{tab:movie_reviews_prompt}. As a result, the two databases contain 11,839 movies, 68,709 users, and 221,242 reviews in total.



\subsection{Persona-augmented User Simulator}~\label{method:user_simulator}
Our user simulator uses GPT-3.5 (\ie, \texttt{GPT-3.5-turbo-1106}; \citet{ouyang2022training}), though in practice, a different model could be used. We explain how we compose persona and how the user simulator generates an utterance.



\paragraph{Persona components.}

We provide the user simulator with persona which includes three types of preferences: general, target, and responsive preferences. For the general preference, we combine three randomly selected review texts of a particular user from the user-review database. This combination represents the user's overall movie taste throughout the dialogue. For the target preference, we select a movie title and its corresponding review text that the user has rated highly (at least 8 out of 10). The user simulator's role is to consistently express the specific attributes of the movie, so that the recommender simulator can eventually suggest the target movie to the user simulator while having a conversation. Lastly, to facilitate the user simulator to provide realistic feedback when a movie is suggested, we obtain the user's review text of the movie from the user-review database as reference.


\paragraph{Utterance generation.}

The goal of the user simulator is to express its detailed preferences consistent with the persona and eventually get the target item as the final recommendation. Given the task description $D_u$, persona $P$, and the dialogue context $C_u^t = (u_1, r_1, u_2, ..., r_t)$ which consists of utterances up to $t$-th turn where $u_i$ and $r_i$ are utterances from user and recommender respectively, GPT-3.5 generates the next response $u_t$ by following the task description under zero-shot setting. Note that the responsive preference in the persona is dynamically obtained from the user-review database when an item is suggested in the middle of the dialogue.
\subsection{Knowledge-augmented Recommender Simulator}~\label{method:rec_simulator}
To generate utterances of recommender with proper recommendations and explanations, we leverage an LLM and a retriever. The retriever first searches top-$k$ context-relevant items, and the LLM performs a step-by-step reasoning to suggest an item the user would like the most based on the expressed preferences and item knowledge. We use GPT-3.5 as the LLM.


\paragraph{Item retrieval for recommendation.} 
Given the dialogue context $C_r^t = (u_1, r_1, u_2, ..., u_t)$, the retriever in the recommender simulator searches context-relevant items among candidate items. Here, we limit the candidate pool to the items for which the user has written a review, as the user simulator of the next turn should be able to provide detailed feedback. In detail, with \texttt{text-embedding-ada-002}, we compute the similarity between the text embedding of the dialogue context and the text embedding of each item knowledge in the candidate pool, which includes the metadata (\eg, title, genre, director, and cast) and the review text of the item. Finally, the top-$k$ texts of the retrieved item knowledge are given to the LLM as input. In order to prevent the dialogue from diverging, we gradually decrease $k$ and forcefully include the target item knowledge in the retrieved pool from a specific turn we set.

\paragraph{Item recommendation with reasoning.} For a corpus with proper recommendations and sufficient explanations, we utilize an LLM's reasoning ability to recommend the optimal item based on the dialogue context.
Given the task description $D_u$, the retrieved item knowledge $I$, and the dialogue context $C_r^t$, an LLM generates the next recommender response $r_t$. It first infers the user's preference based on the dialogue context. Then, given the retrieved top-$k$ texts of item knowledge, the LLM selects the most preferable item by the user. Lastly, the LLM generates an utterance that recommends the selected item with a plausible explanation. The prompt is shown in Table \ref{tab:RG_prompt}.


\subsection{Dialogue Generation}~\label{method:gen_dialogues}
Figure \ref{fig:method} shows the architecture for the interaction between the user and recommender simulators. We first initialize the dialogue context with an utterance pair (\ie, two dialogue turns), and then each simulator starts interacting based on the given utterances. To simulate the real-world recommendation conversation, we provide different inputs in each turn. On the user simulator side, we provide a persona to generate utterances with specific preferences. On the recommender simulator side, we use the retriever to find suitable items for the dialogue context. Then the LLM reasons about the user's preference and chooses the most proper item for the user and recommends the item along with a detailed explanation based on the item knowledge.

{\renewcommand{\arraystretch}{1.35}
    \begin{table}[t] \begin{center}
    \small
    \setlength{\tabcolsep}{3pt}
    \begin{tabularx}{\linewidth}{X}
        \toprule
        \textbf{Seeker:} Hi there! I'm in the mood to watch a movie. Can you recommend something? \\
        \textbf{Recommender:} Absolutely! What kind of movie are you in the mood for? Any specific genre or theme? \\
        \textbf{Seeker:} I'm generally a fan of movies with strong performances by the cast and enthralling cinematic accomplishments. I also appreciate well-written plots and screenplays. \\
        \textbf{Recommender:} I think you would really enjoy `Darkest Hour (2017)'. It has a strong performance by Gary Oldman as Winston Churchill and the attention to detail in bringing the period to life is enthralling. Plus, the focus on Churchill's point of view and how he came to power makes for a compelling and well-written plot. \\
        \textbf{Seeker:} I can see how `Darkest Hour' fits some of my preferences, but I'm really in the mood for something with more action and combat. I also enjoy movies with historical accuracy and realistic depictions of challenges faced by characters in combat situations. Do you have anything that fits those criteria? \\
        \textbf{Recommender:} How about watching `Godzilla (1998)'? It has great action sequences involving Godzilla, and the movie stays faithful to the original 1950's movie concept. The special effects and pacing are also very well done, and the film balances action, sci-fi, and thriller elements. Plus, it's based on historical events and provides a realistic depiction of characters facing challenges in combat situations. \\
        \textbf{Seeker:} I appreciate the recommendation, but `Godzilla (1998)' isn't quite what I'm in the mood for right now. I'm really looking for something with a focus on the life of an Army crew and the challenges they face in combat. Do you have anything else in mind that fits that description? \\
        \textbf{Recommender:} I think `Fury (2014)' would be a perfect fit for what you're looking for. It provides a realistic depiction of the challenges faced by an Army Sherman tank crew in combat situations. The performances by the cast are strong, and the screenplay is well-written, delivering a compelling and enthralling cinematic experience. \\
        \textbf{Seeker:} That sounds like exactly what I'm looking for! I'll definitely give `Fury (2014)' a watch. Thanks for the recommendation! \\
        \bottomrule
    \end{tabularx}
    \vspace{-3pt}
    \caption{
        Turn-by-turn dialogue sample from \ourdata{}. 
    }
    \vspace{-10pt}
    \label{tab:example}
\end{center}\end{table}}



\subsection{Dialogue Filtering}
\label{method:filtering}
\paragraph{Basic filtering.} 
Starting with an initial set of dialogues sampled from GPT-3.5, we remove dialogues (1) where utterances are generated repeatedly, (2) where the user simulator explicitly mentions the target item, or (3) where the user accepts other items instead of the target item. 
Through these data filtering procedures, the gathered dialogues could be better utilized to investigate the conversational recommendation methods.

\paragraph{Preference-based filtering.} 
We also filter out dialogues based on user preference to maintain dialogue consistency and enhance preference-based recommendations. 
Specifically, we (1) compute the natural language inference (NLI) score between the persona and user utterances, filtering out the dialogues that include any utterance that contradicts the persona, and (2) extract the recommender simulator's guess of user preference to check for contradictions with user utterances. 
If there is any contradicted utterance, we consider that the recommender simulator fails to model the user's preference and discard such dialogues.

\paragraph{Final dataset.} By applying a series of dialogue filtering, we obtain \ourdata{} with 57.2K dialogues with more than 4k users and 9k items, where 22.5\% of the initial dialogues are removed.

\section{Experiments}

\subsection{Evaluation on Dataset Quality}
\paragraph{Datasets.} We conduct experiments on ReDial \citep{li2018towards} and INSPIRED \citep{hayati2020inspired}. ReDial is an English CRS dataset about movie recommendations, and is constructed through crowd-sourcing workers on Amazon Mechanical Turk (AMT). Similar to ReDial, INSPIRED is also an English CRS dataset about movie recommendations, but with a smaller size. These two datasets are widely used for evaluating CRS models. 

\paragraph{Human evaluation.} To assess the relative quality of \ourdata{} compared to previous datasets, we conduct human evaluation through head-to-head comparison on Amazon Mechanical Turk, comparing \ourdata{} with two widely used open-domain dialogue datasets: ReDial \citep{li2018towards} and INSPIRED \citep{hayati2020inspired}. We randomly sample 100 dialogues from each dataset and evaluate them according to six criteria: (1) user-control, (2) expertise, (3) specificity of user preference, (4) relevance, (5) flow naturalness, and (6) consistency. Judges are asked to select a better one between two given dialogues, regarding each criterion. Further details are in Appendix \ref{app:human_eval}.

\begin{figure}[t!]
    \centering
    \includegraphics[width=\linewidth]{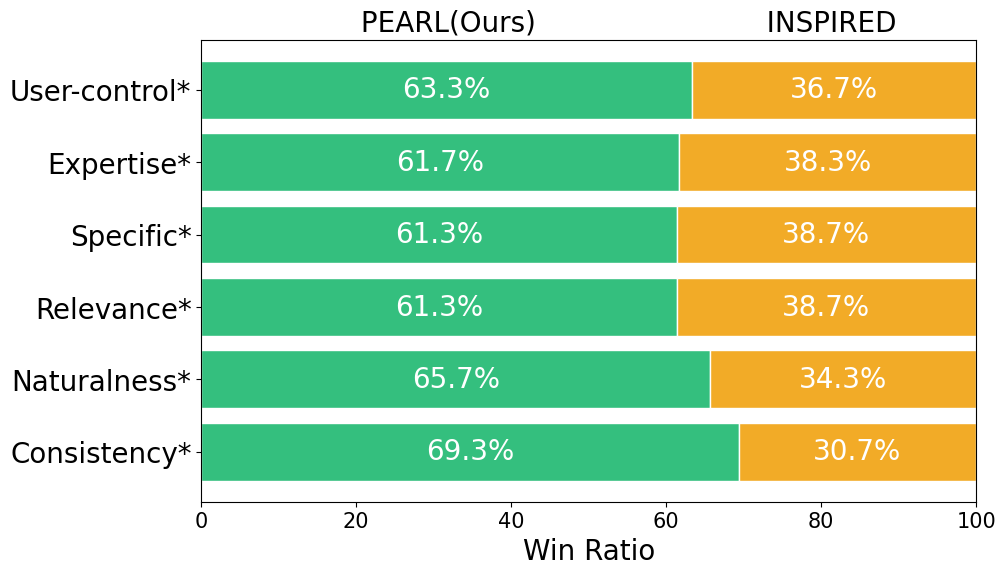}
    \caption{Results of human evaluation on head-to-head comparison between conversations sampled from \ourdata{} and those from INSPIRED. (*: p-value < 0.05)}
    \label{fig:human_eval_inspired}
\end{figure}

Figure \ref{fig:human_eval_redial} and \ref{fig:human_eval_inspired} summarizes the head-to-head comparison of \ourdata{} and human-annotated datasets. Despite being fully machine-generated, human raters judge \ourdata{} as better in quality compared to both ReDial and INSPIRED.

\begin{figure}[t!]
    \centering
    \includegraphics[width=\linewidth]{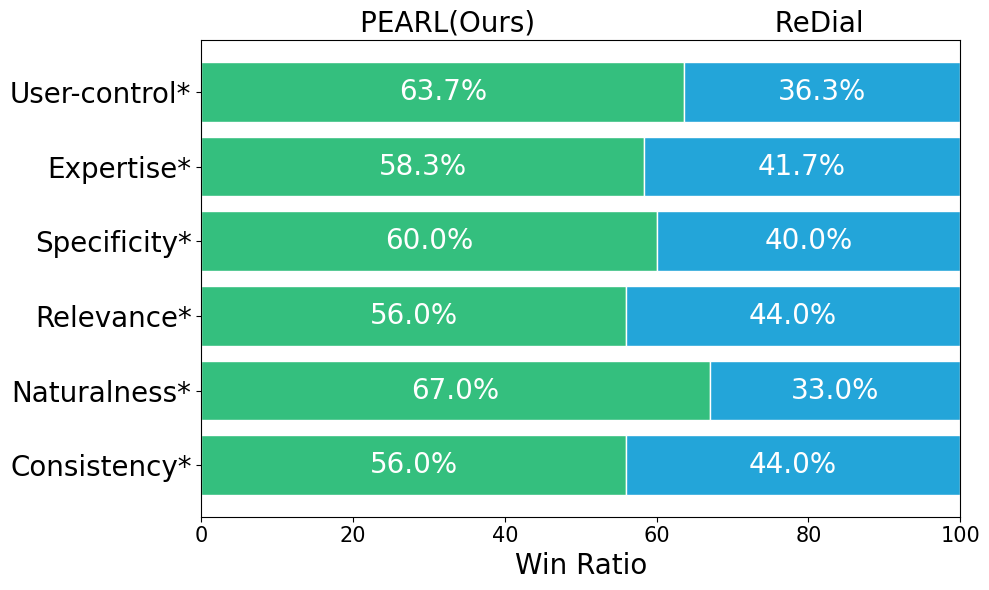}
    \caption{Results of human evaluation on head-to-head comparison between conversations sampled from \ourdata{} and those from ReDial. (*: p-value < 0.05)}
    \label{fig:human_eval_redial}
\end{figure}

{\renewcommand{\arraystretch}{1.2}
    \begin{table}[t!] \begin{center}
    \begin{adjustbox}{width=\columnwidth}
    \begin{tabular}{lccc}
        \toprule
         & ReDial & INSPIRED & \textbf{\ourdata{}} \\
        \midrule  
        \# of dialogues  & 10,006 & 1,001 & 57,277 \\ 
        \# of utterances  & 182,150 & 35,811 & 548,061 \\ 
        2-gram specificity & 65.44 & 119.56 & 141.79 \\
        3-gram specificity & 65.97 & 123.01 & 149.75 \\
        4-gram specificity & 65.37 & 122.81 & 153.00 \\
        \bottomrule
    \end{tabular}
    \end{adjustbox}
    \caption{
        Statistics of \ourdata{} compared to ReDial and INSPIRED. The table shows the number of dialogues, utterances, and n-gram specificities for each dataset.
    }
    \vspace{-10pt}
    \label{tab:dataset_stats}
\end{center}\end{table}}

\begin{table}[t!]
    \footnotesize 
    \centering
    \begin{adjustbox}{width=0.8\columnwidth} 
        \begin{tabular}{lccc}
            \toprule
            \multicolumn{4}{c}{Inter-dialogue similarity of user utterances} \\
            \midrule  
            Ours & \multicolumn{3}{r}{0.1900} \\
            w/o Persona & \multicolumn{3}{r}{0.1962} \\
            \bottomrule
        \end{tabular}
    \end{adjustbox}
    \caption{Inter-dialogue similarity of \ourdata{} and ablated dialogues generated without persona.}
    \vspace{-5pt} 
    \label{tab:ablation_study_persona}
\end{table}


\paragraph{User preference analysis.} 
We compare the specificity of user preferences across diverse CRS datasets in Table \ref{tab:dataset_stats} as the capability to understand specific user preferences is crucial in suggesting personalized and satisfactory recommendations. We measure it by concatenating all user utterances (\ie, preference) of a dialogue and obtaining the number of unique $n$-grams within it. According to Table \ref{tab:dataset_stats}, \ourdata{} contains more unique expressions than ReDial \citep{li2018towards} or INSPIRED \citep{hayati2020inspired}, indicating \ourdata{} has less generic and more specific user preferences.

To investigate the effect of utilizing persona in the data synthesis process, we additionally conduct an ablation study on how persona-augmentation yields more diverse and distinct preferences explicitly through user utterances. By comparing the inter-dialogue similarity (\ie, semantic similarity between the concatenated user utterances from two arbitrary dialogues) in Table \ref{tab:ablation_study_persona}, we observe that synthesizing a dialogue with persona input yields utterances with more distinct preferences that are less similar to each other.




\begin{table}[t!]
    \centering
    
    \begin{adjustbox}{width=0.8\columnwidth} 
        \begin{tabular}{l c c c}
            \toprule
             & ReDial & INSPIRED & \ourdata{} \\
            \midrule
            \# of words & 11.01 & 14.62 & 38.81 \\
            \bottomrule
        \end{tabular}
    \end{adjustbox}
    \caption{Average number of words per recommender utterance in ReDial, INSPIRED, and \ourdata{}.}
    \vspace{-5pt} 
    \label{tab:ablation_study_knowledge}
\end{table}

{\renewcommand{\arraystretch}{1.35}
    \begin{table}[t] \begin{center}
    \small
    \setlength{\tabcolsep}{3pt}
    \begin{tabularx}{\linewidth}{X}
        \toprule
        \textbf{ReDial}: You might like "The Boss Baby (2017)" that was a good movie. \\
        \midrule
        \textbf{INSPIRED}: Have you seen the movie Hustlers yet? It is a little bit of a mix of comedy, drama and thriller. \\
        \midrule
        \textbf{\ourdata{}}: How about "The Addams Family (1991)"? It's a dark comedy with supernatural elements and a great cast, including standout performances from Raul Julia and Christopher Lloyd. Plus, it has a macabre and humorous vibe that I think you'll enjoy.\\
        \bottomrule
    \end{tabularx}
    \vspace{-3pt}
    \caption{Sample utterances from ReDial, INSPIRED, and \ourdata{}.}
    \vspace{-10pt}
    \label{tab:knowledge_ab_example}
\end{center}\end{table}}

\paragraph{Knowledge-augmented recommendation analysis.} 
We compare the degree of richness in explanations of recommender utterances of ReDial, INSPIRED, and \ourdata{} in Table \ref{tab:ablation_study_knowledge} as it is an important factor for knowledgeable and explainable conversational recommender systems. 
To quantitatively measure the richness, we calculate number of words used in a single recommender utterance in average. For ReDial which contains several consecutive recommender utterances, we combine them into one utterance as we calculate. Also, we show qualitative examples of each dataset in Table \ref{tab:knowledge_ab_example}. While the utterances of Redial and INSPIRED are relatively shallow and brief, the utterance of \ourdata{} explains about the item in a great detail which may enhance user satisfaction.



%



\paragraph{Data scale comparison.} With 57,277 dialogues, \ourdata{} is the largest in scale compared to existing crowdsourced conversational recommendation datasets (Table \ref{tab:dataset_stats}). It contains more than 500k utterances, each reflecting the preferences of real-world users, thereby providing a rich resource for training conversational recommender systems.

\paragraph{Cost \& time-efficiency comparison.} 
Synthesizing \ourdata{} by utilizing the simulators grounded on persona and knowledge is significantly more efficient than traditional dialogue crowdsourcing datasets in both cost and time. For instance, INSPIRED dataset took four months to crowdsource a total of 1,001 dialogues. 
In contrast, our data generation process with GPT-3.5-turbo-1106 generates more than 57k dialogue datasets in just one week. Furthermore, in terms of cost, the INSPIRED dataset costs an average of \$5 per dialogue, whereas the data synthesis process incurs a total cost of about \$0.02 per dialogue.

\begin{figure}[t!]
    \centering
    \includegraphics[width=\linewidth]{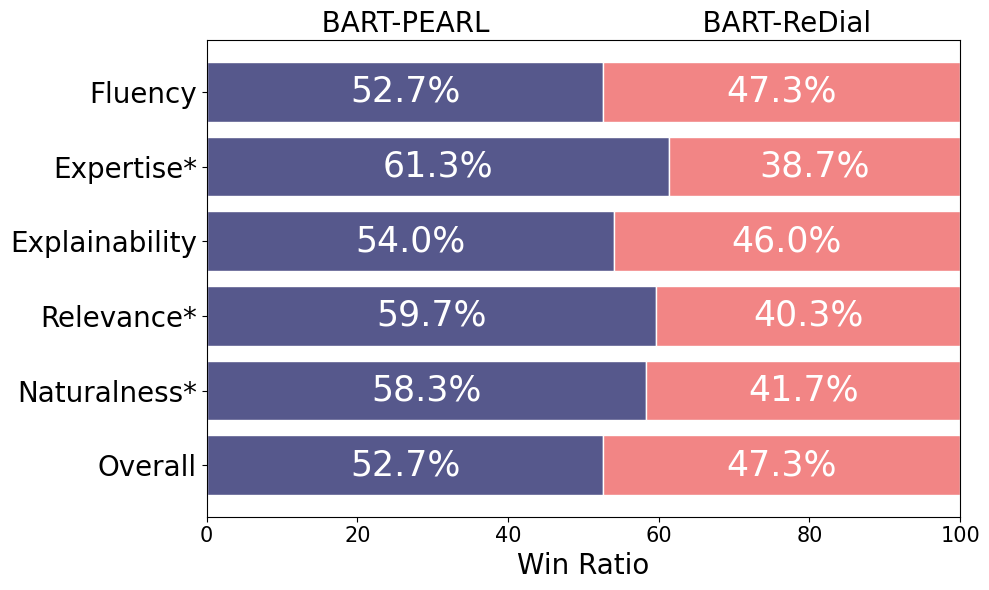}
    \caption{Results of head-to-head comparison human evaluation between responses generated from BART trained on \ourdata{} and on ReDial. (*: p-value < 0.05)}
    \label{fig:human_eval_response_quality}
\end{figure}

\subsection{Evaluation on Dataset Utility}~\label{experiments:utility}

\paragraph{Human evaluation.}

To qualitatively assess the utility of \ourdata{}, we perform human evaluation that compares the responses of BART \citep{lewis-etal-2020-bart} trained on ReDial (BART-ReDial) and on \ourdata{} (BART-\ourdata{}) given the same dialogue context from INSPIRED, which is an unseen dataset for the models. 
We sample 100 dialogue contexts randomly from INSPIRED test set and ask three human judges per each dialogue context to select a better response between the two in terms of six distinct criteria: (1) fluency, (2) expertise, (3) explainability, (4) relevance, (5) naturalness, and (6) overall. Further details are in Appendix \ref{app:human_eval}.

 Although \ourdata{} is the only machine-generated dataset, Figure \ref{fig:human_eval_response_quality} shows that BART-\ourdata{} consistently outperforms BART-ReDial across all evaluation criteria. Specifically, BART-\ourdata{} shows the largest gap in terms of \textit{expertise}, highlighting the utility of our approach in enriching recommender responses with a deeper understanding and insight into the movie domain.


\paragraph{Automatic evaluation.} We conduct experiments on response generation and recommendation tasks to assess the utility of \ourdata{}, following previous CRS works \citep{zhou2020improving,wang2022towards}. For the response generation task, we employ diverse language models such as BART \citep{lewis-etal-2020-bart}, UniCRS \citep{wang2022towards}, and PECRS \citep{ravaut-etal-2024-parameter} to evaluate how effectively \ourdata{} enhances the diversity of the outputs. We evaluate the models using context from an unseen dataset, INSPIRED, and adopt Distinct-$n$ ($n$=3,4) as the metric. For the recommendation task, BERT \citep{devlin-etal-2019-bert}, UniCRS, and PECRS are employed. We use Recall@$k$ ($k$=1, 10, 50) as the metric, which indicates the percentage of target items correctly identified within the top-$k$ recommendations. In addition to the downstream models, we also provide performances for zero-shot GPT-3.5 (\texttt{GPT-3.5-turbo-1106}) on both tasks to assess the capabilities of large language models in handling these tasks without task-specific fine-tuning.\footnote{GPT-3.5 is evaluated in zero-shot setting on recommendation and response generation tasks following \citet{he2023large}.}



\begin{table}
    \centering
    \small
    \begin{tabular}{ l c c} 
        \toprule[1.0pt]
        Model  & Dist-3 & Dist-4  \\ 
        \toprule[1.0pt] 
        BART-ReDial & 0.6220 & 0.5057 \\
        BART-\ourdata{} & \textbf{0.9241} & \textbf{0.8861} \\
        \midrule
        UniCRS-ReDial & 0.5413 & 0.3667 \\
        UniCRS-\ourdata{} & \textbf{0.9338} & \textbf{0.9007} \\
        \midrule
        PECRS-ReDial & 0.6798 & 0.5906 \\
        PECRS-\ourdata{} & \textbf{0.9132} & \textbf{0.8947} \\
        \midrule
        GPT-3.5     & 0.9256 & 0.8910 \\
        \bottomrule[1.0pt]
    \end{tabular}
    \caption{Response generation performances on INSPIRED. }
    \label{tab:main_table_con}
\end{table}

\begin{table}
    \centering
    \small
    \begin{tabular}{ l c c c } 
        \toprule[1.0pt]
        Model & R@1  & R@10 & R@50   \\ 
        \toprule[1.0pt] 
        BERT-\ourdata{} & 0.0018 & 0.0208 & 0.0736 \\
        UniCRS-\ourdata{} & \textbf{0.0310} & \textbf{0.0697} & \textbf{0.1202} \\
        PECRS-\ourdata{} & 0.0151 & 0.0339 & 0.0798 \\
        \midrule
        GPT-3.5  & 0.0071 & 0.0355 & 0.0709 \\
        \bottomrule[1.0pt]
    \end{tabular}
    \caption{Recommendation performances on \ourdata{}.}
    \label{tab:main_table_rec}
\end{table}


Table \ref{tab:main_table_con} compares the response generation performances of models trained on \ourdata{} and on ReDial, when evaluated on INPSIRED. The results show the effectiveness of \ourdata{} in improving the diversity of responses generated by downstream models. Moreover, GPT-3.5 shows lower performances than UniCRS-\ourdata{} and PECRS-\ourdata{}, despite GPT-3.5's much larger model size, indicating \ourdata{} is challenging and of high-quality.

Table \ref{tab:main_table_rec} presents the recommendation performances of models trained and evaluated on \ourdata{}. As the recommendation ability of a trained model highly depends on the distribution of the training data, we do not evaluate with an unseen dataset as done in response generation experiments. We find all models exhibit low performances, indicating that \ourdata{} presents a significant challenge. Interestingly, GPT-3.5 shows a poor performance, suggesting that although \ourdata{} is generated by the same model (\ie, GPT-3.5), there is no inherent advantage for the model. This observation underscores the need for future research to focus on developing models specifically tailored to \ourdata{}, which leverages real-world data rather than relying on parametric knowledge.

{\renewcommand{\arraystretch}{1.35}
    \begin{table}[t] \begin{center}
    \small
    \setlength{\tabcolsep}{3pt}
    \begin{tabularx}{\linewidth}{X}
        \toprule
        \textbf{\textit{Dialogue context}} \\
        ... \\
        \textbf{Seeker}: I accept this recommendation and would love to watch it with my family! Thank you for your help. \\
        \textbf{Recommender}: You're very welcome! I hope you enjoy it! \\
        \textbf{Seeker}: If we finish this movie early, do you have children's Christmas movie to recommend? So far our favorite is Elf. \\
        \midrule
        \textbf{\textit{Model responses}} \\
        \textbf{GPT-3.5}: I would recommend the movie "The Polar Express" as a great children's Christmas movie. It's a heartwarming and magical adventure that is perfect for the holiday season. Enjoy watching it with your family! \\
        \textbf{BART-ReDial}: Yes, I love Elf! It's a great family movie! \\
        \textbf{BART-\ourdata{} (Ours)}: How about watching `The Lion King'? It's a classic animated film with a heartwarming story and memorable characters. It's perfect for family viewing and has a great message about family and friendship. \\
        \midrule
        \textbf{Ground-truth}: I would absolutely recommend Rise of the Guardians it's a kid's movie, but definitely funny so adults can enjoy too. \\
        \bottomrule
    \end{tabularx}
    \vspace{-3pt}
    \caption{
        Cherry-picked sample responses from BART-ReDial and BART-\ourdata{} to a context in INSPIRED. The full example is in Table \ref{tab:case_study_cherry}.
    }
    \vspace{-10pt}
    \label{tab:case_study_cherry_short}
\end{center}\end{table}}

{\renewcommand{\arraystretch}{1.35}
    \begin{table}[t] \begin{center}
    \small
    \setlength{\tabcolsep}{3pt}
    \begin{tabularx}{\linewidth}{X}
        \toprule
        \textbf{\textit{Dialogue context}} \\
        ... \\
        \textbf{Seeker}: I like both, but I'm looking for something super hero related, like Guardians of the Galaxy.  \\
        \textbf{Recommender}: Oh! I have a few ideas of what you have in mind. Are comic book characters interesting at all?  \\
        \textbf{Seeker}: Yes! I love comic book characters, like Spider man, Ironman and Captain America.  \\
        \textbf{Recommender}: Okay. Would you like to watch this trailer for the Avengers?  \\
        \textbf{Seeker}: Oh that would be great!  \\
        \midrule
        \textbf{\textit{Model responses}} \\
        \textbf{GPT-3.5}: I would recommend you watch the trailer for "Avengers: Endgame." It features all your favorite comic book characters like Iron Man, Captain America, and Spider-Man, and it's a super hero-related movie just like Guardians of the Galaxy. Enjoy watching the trailer! \\
        \textbf{BART-ReDial}: Okay. I would also recommend `The Avengers (2012)' \\
        \textbf{BART-\ourdata{} (Ours)}: How about `The Avengers (2012)'? It's an action-packed superhero movie with a great cast and some really interesting characters. \\
        \midrule
        \textbf{Ground-truth}: I haven't actually watched it yet. It is the one with that Thanos character and the Justice league, right? \\
        \bottomrule
    \end{tabularx}
    \vspace{-3pt}
    \caption{
        Lemon-picked sample responses from BART-ReDial and BART-\ourdata{} to a context in INSPIRED. The full example is in Table \ref{tab:case_study_lemon}.
    }
    \vspace{-10pt}
    \label{tab:case_study_lemon_short}
\end{center}\end{table}}

\subsection{Case Study}

Table \ref{tab:case_study_cherry_short} and \ref{tab:case_study_lemon_short} present cherry-picked and lemon-picked responses from BART-ReDial and BART-\ourdata{}, respectively. BART-\ourdata{} consistently provides explanations that elucidate what the recommended item is and why the recommender suggests the item, while BART-ReDial offers shorter responses without such details. In the lemon-picked example, BART-\ourdata{} seems to forgot the previous mention of \textit{``The Avengers (2012)''}. However, it still manages to explain the recommended item, demonstrating the model's explainability trained on our synthesized dataset.

\section{Conclusion}
In this work, we introduce \ourdata{}, a large-scale conversational recommendation dataset constructed using LLM simulators augmented with persona or knowledge from real-world reviews. Our comprehensive experiments validate \ourdata{}'s superior quality and utility in training models compared to existing datasets.

\ourdata{} paves the way for future research in developing effective conversational recommender systems, particularly those emphasizing explainability, knowledge retrieval, and reasoning abilities. Also, \ourdata{} offers a practical opportunity for developing small and deployable systems capable of handling specific user feedback and providing satisfactory recommendations.
\section*{Limitations}
As we generate recommendation dialogues using LLM-based simulators, the choice of a language model (\ie, GPT-3.5 in this work) will impact the quality of dialogue created. One of the possible future directions may include curating recommendation dialogues by using simulators based on different language models and investigating the difference between generated dialogues and utterances.

\section*{Ethical Considerations}
Our work utilizes a large language model and real-world reviews for recommendation dialogue generation and filters out generated dialogues in terms of user preference. As our filtering mechanism does not address considerations related to dialogue safety, the users who employ our data generation process should be mindful of this limitation and consider the incorporation of additional filtering steps to mitigate potential biases or toxic content. We ensure that workers hired through Amazon Mechanical Turk receive fair compensation. We offer an effective hourly rate exceeding \$15, based on the estimated time required to complete the tasks.

\section*{Acknowledgements}
This work was supported by the IITP grant funded by the Korea government (MSIT) (No. RS-2020-II201361) and the NRF grant funded by the Korea government (MSIT) (No. RS-2023-00244689). We thank Hyunseo Kim and Soyeon Chun for their participation in experiments

%

\bibliography{custom}
\newpage
\appendix

\section{Implementation Details}
\label{app:implement_details}

\subsection{Large language model}
\label{ssec:llm}
In this work, we utilize GPT-3.5 (GPT-3.5-turbo-1106) for our user simulator and recommender simulator. GPT-3.5 is an LLM based on InstructGPT~\citep{ouyang2022training}.\footnote{https://openai.com/blog/chatgpt} 
The prompts used to generate persona, item knowledge, seeker's utterance and recommender's response are in Table~\ref{tab:user_review_prompt}, Table~\ref{tab:movie_reviews_prompt}, Table~\ref{tab:seeker_prompt}, and Table~\ref{tab:RG_prompt}, respectively.

\subsection{Computational Resources and API Cost}
We run the BART and NLI models on eight NVIDIA RTX A5000 GPUs. For ChatGPT API usage, we use about 0.02 dollars per one dialogue generation.

\subsection{Natural language inference model}
We leverage an external natural language inference (NLI) model in the dialogue filtering process to obtain higher-quality recommendation conversation data. In particular, in preference-based filtering, we filter it out if the NLI model predicts the logical relationship between persona and user's utterance is \textit{contradiction} with $\delta > 0.7$. As for the NLI model, we use RoBERTa~\cite{liu2019roberta} fine-tuned on the DNLI dataset~\cite{welleck-etal-2019-dialogue}.

\subsection{Downstream models}
We utilize source codes from \url{https://github.com/RUCAIBox/UniCRS} for UniCRS and \url{https://github.com/Ravoxsg/efficient_unified_crs} for PECRS experiments. For BERT, we take a naive method to predict the most plausible movie based on the dialogue context. For BART, we also take a general setting of sequence-to-sequence modeling to generate the next response when given the dialogue context.

\section{Human Evaluation Metrics}
\label{app:human_eval}
\subsection{Dataset Quality}
\begin{figure*}
    \centering
    \includegraphics[width=1\linewidth]{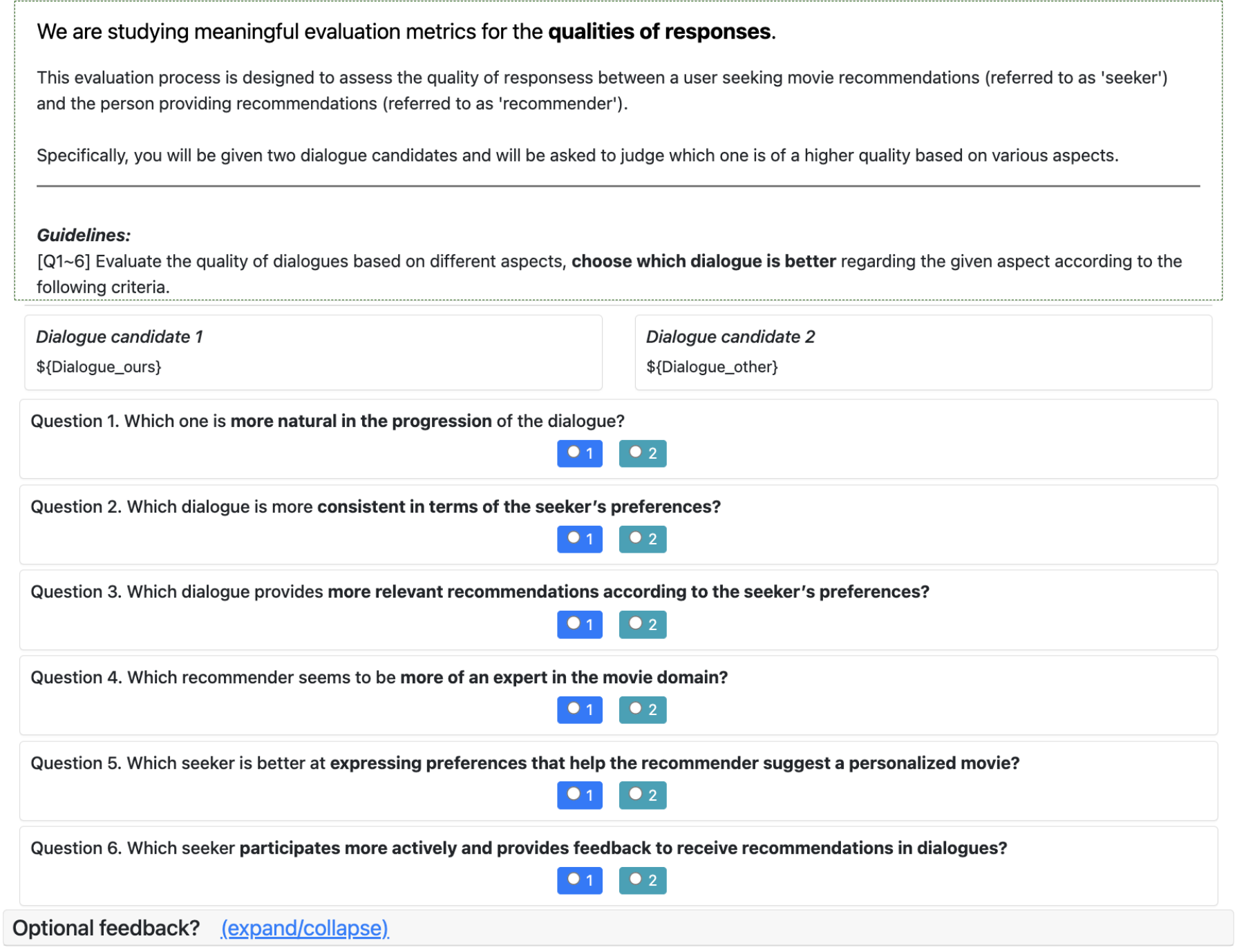}
    \caption{Interface for human evaluation on dataset quality.}
    \label{fig:rg_human_amt_quality}
\end{figure*}

We outsource a human evaluation comparing our \ourdata{} and previous datasets via Amazon Mechanical Turk (AMT).
We show the interface for the evaluation in Figure~\ref{fig:rg_human_amt_quality}. 
We ask the human judges to compare the dialogues based on the following criteria:
\begin{itemize}
    \item \textbf{User-control}: Which seeker participates more actively and provides feedback to receive recommendations in dialogues?
    \item \textbf{Expertise}: Which recommender seems to be more of an expert in the movie domain?
    \item \textbf{Specificity}: Which seeker is better at expressing preferences that help the recommender suggest a personalized movie?
    \item \textbf{Relevance}: Which dialogue provides more relevant recommendations according to the seeker's preferences?
    \item \textbf{Flow Naturalness}: Which one is more natural in the progression of the dialogue?
    \item \textbf{Consistency}: Which dialogue is more consistent in terms of the seeker's preferences?
 
\end{itemize}

During each stage of voting, human judges are given two dialogue candidates and asked to judge which one is of a higher quality based on the above criteria. 

\subsection{Dataset Utility}
\begin{figure*}
    \centering
    \includegraphics[width=1\linewidth]{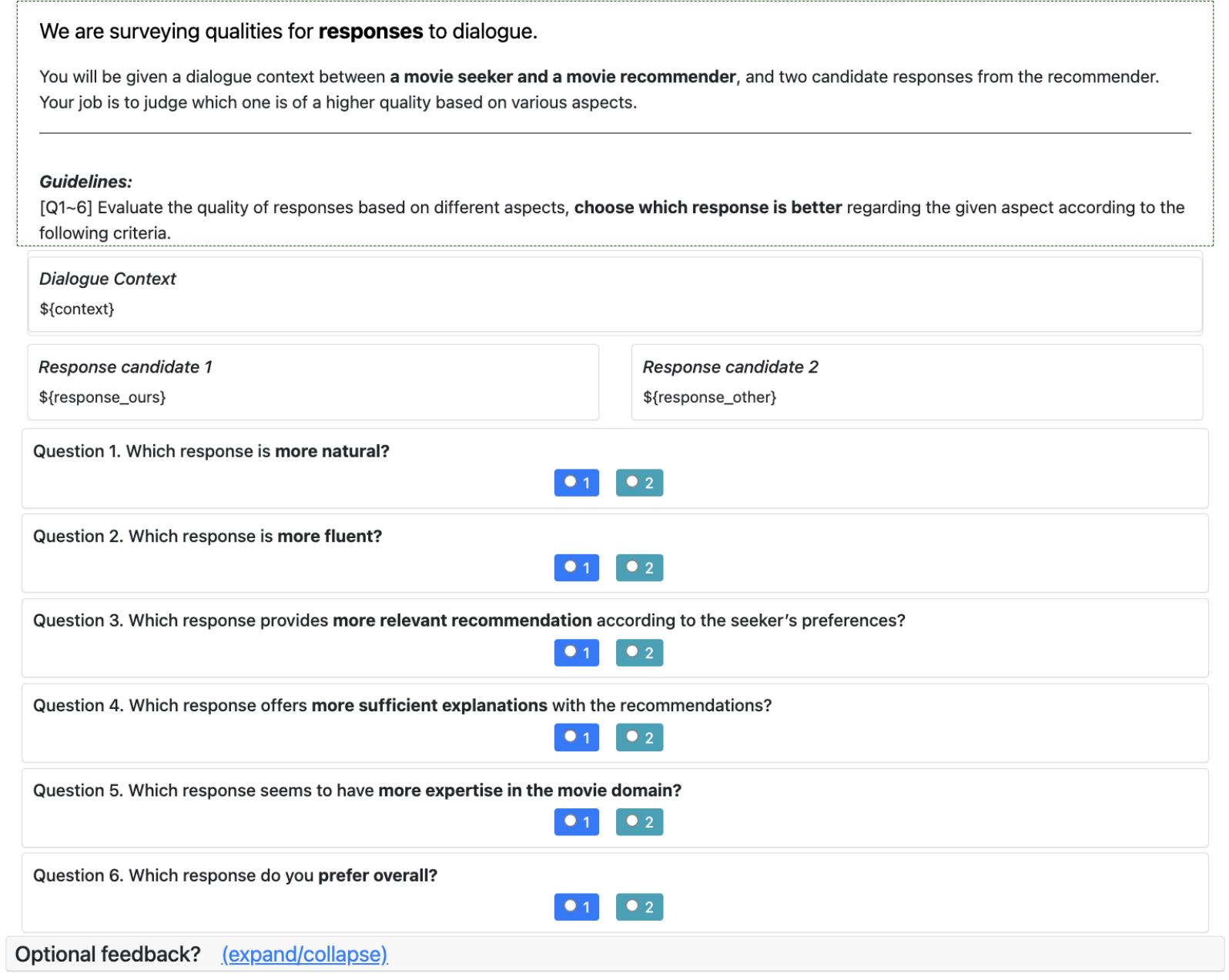}
    \caption{Interface for human evaluation on dialogue utility.}
    \label{fig:rg_human_amt_utility}
\end{figure*}

We also outsource a human evaluation comparing the responses of the BART model trained on \ourdata{} and ReDial dataset.
We show the interface for this evaluation in Figure~\ref{fig:rg_human_amt_utility}.

We ask the human judges to compare the responses and select the better one based on the following criteria: 
\begin{itemize}
    \item \textbf{Fluency}: Which response is more fluent?
    \item \textbf{Expertise}: Which response seems to have more expertise in the movie domain?
    \item \textbf{Explainability}: Which response offers more sufficient explanations with the recommendations?
    \item \textbf{Relevance}: Which response provides more relevant recommendation according to the seeker's preference?
    \item \textbf{Naturalness}: Which response is more natural?
    \item \textbf{Overall}: Which response do you prefer overall?
    
\end{itemize}

\begin{figure*}[t!]
    \centering
    \includegraphics[width=0.5\linewidth]{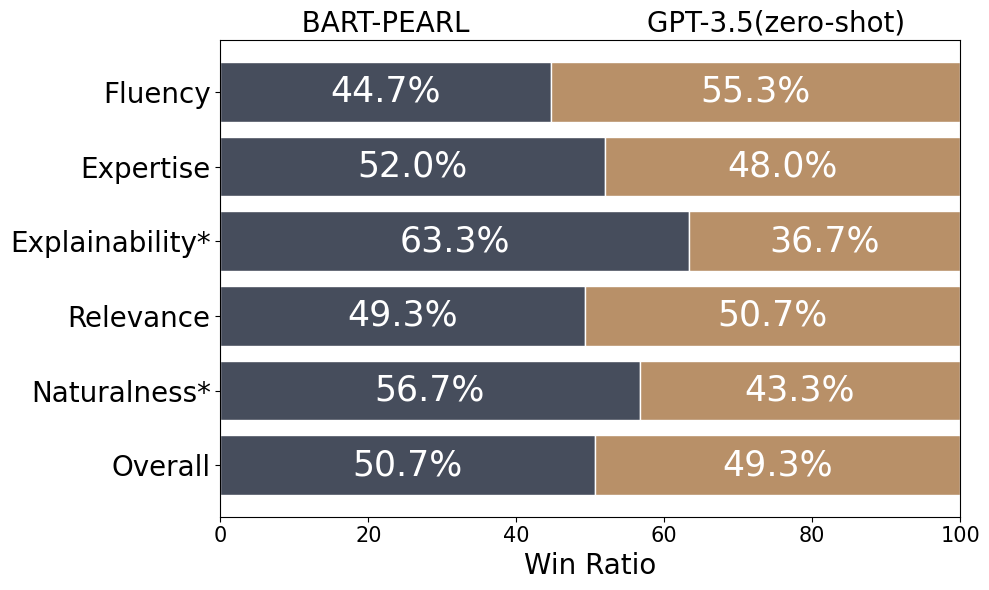}
    \caption{Results of head-to-head comparison human evaluation between responses generated from BART trained on \ourdata{} and from GPT-3.5 zero-shot inference. (*: p-value < 0.05)}
    \label{fig:human_eval_bart_gpt}
\end{figure*}

\begin{table*}
    \centering
    \small
    \begin{tabular}{ l c c c c c c} 
        \toprule[1.0pt]
        Model & ROUGE-1 & ROUGE-2 & Dist-1 & Dist-2 & Dist-3 & Dist-4 \\ 
        \toprule[1.0pt] 
        BART-ReDial & 0.1370 & \textbf{0.0264} & \textbf{0.9826} & 0.7596 & 0.6208 & 0.4964 \\
        BART-\ourdata{} & \textbf{0.1642} & 0.0241 & 0.8803 & \textbf{0.9594} & \textbf{0.9365} & \textbf{0.9047} \\
        \midrule
        UniCRS-ReDial & 0.0925 & 0.0097 & \textbf{0.9545} & 0.7916 & 0.6288 & 0.4635 \\
        UniCRS-\ourdata{} & \textbf{0.2153} & \textbf{0.0218} & 0.7551 & \textbf{0.8997} & \textbf{0.9352} & \textbf{0.9027} \\
        \midrule
        PECRS-ReDial & 0.1979 & \textbf{0.0573} & \textbf{0.7995} & 0.7649 & 0.6801 & 0.6039 \\
        PECRS-\ourdata{} & \textbf{0.2619} & 0.0497 & 0.7637 & \textbf{0.9058} & \textbf{0.9182} & \textbf{0.9074} \\
        \midrule
        GPT-3.5 & 0.2256 & 0.0330 & 0.8990 & 0.9620 & 0.9354 & 0.9046 \\
        \bottomrule[1.0pt]
    \end{tabular}
    \caption{Response generation performances on E-ReDial \citep{guo2023towards} \textit{Test-Rec} subset.}
    \label{tab:eredial_table}
\end{table*}

\newpage
\section{Additional Experimental Results}
\paragraph{Dataset utility for practical use}
Our research aims to tailor a model specific to the conversational recommendation task, by utilizing our proposed dataset. While a large language model (\eg GPT-3.5) may seem plausible as a conversational recommender due to its capability to generate human-like text, \ourdata{} and its downstream models are still necessary. First of all, a large language model (\eg GPT-3.5) is practically inappropriate to be deployed in real-time services due to its latency, cost, and bias issues. Also, while constructing the dialogues in \ourdata{}, we have taken into account the relationship between the actual user history and the the target recommendation. We believe that large language models lack in such collaborative interaction knowledge, as shown in Table \ref{tab:main_table_rec} where it achieves only 0.0355 in Recall@10 in our test set.

In addition, we ran a human evaluation between responses of GPT-3.5 and BART trained on \ourdata{}(BART-\ourdata{}), given INSPIRED contexts. We randomly sampled 50 responses from each model and used the same criteria as dataset utility human evaluation in the paper. As shown in Figure \ref{fig:human_eval_bart_gpt}, responses from BART-\ourdata{} are generally preferred to those from GPT-3.5. Although BART shows inferior fluency and decent relevance, it outperforms GPT-3.5 in expertise and explainability, the two most important aspects emphasized throughout our paper. We believe that augmenting responses with domain-specific knowledge in \ourdata{} helped the downstream model enrich explanations beyond the basic information of the movie.

\paragraph{Explainability of downstream models}
Table \ref{tab:eredial_table} shows experimental results for downstream models (BART, UniCRS, and PECRS) evaluated on E-ReDial \citep{guo2023towards}. E-ReDial consists of high-quality explanations manually annotated by human workers, based on ReDial dialogues. To asses the models' ability in providing explanations for recommendations, we utilize the \textit{Test-Rec} subset of the test set, which always contains recommendations and explanations as suggested by the authors.

According to Table \ref{tab:eredial_table}, models trained on \ourdata{} outperforms those trained on ReDial on most metrics. For the reference-based metrics (ROUGE-1, ROUGE-2) where they underperform, we speculate that models trained on ReDial benefit from sharing the recommendation distribution with E-ReDial.





\begin{table*}
    \small
    \centering
    \begin{tabular}{p{14cm}}
    \toprule
    \textbf{Prompt} \\
    \midrule
    Given a review written by you, summarize what you liked and disliked about the movie, under [Like] and [Dislike] respectively. If there is nothing to mention about like/dislike, simply write "None." under the corresponding tag.
    \\
    \\
    Here is the movie review written by you: \\ \{review\} \\
     \bottomrule
    \end{tabular}
    \caption{The prompt for summarizing a review of a user.}
    \label{tab:user_review_prompt}
\end{table*}

\begin{table*}
    \small
    \centering
    \begin{tabular}{p{14cm}}
    \toprule
    \textbf{Prompt} \\
    \midrule
    Given some popular reviews about \{movie title\}, describe what people liked and disliked about the movie, under [Like] and [Dislike] respectively. If there is nothing to mention about like/dislike, simply write "None." under the corresponding tag.
    \\ \\
    Here are some basic information about the movie and reviews about it: \\ 
    Movie Title: \{movie title\} \\
    Genre: \{genre\} \\
    Director: \{director\} \\
    Cast: \{cast\} \\
    Reviews: \\
    \{reviews\} \\
     \bottomrule
    \end{tabular}
    \caption{The prompt for summarizing some popular reviews of a movie.}
    \label{tab:movie_reviews_prompt}
\end{table*}
\begin{table*}
    \small
    \centering
    \begin{tabular}{p{14cm}}
    \toprule
    \textbf{Prompt} \\
    \midrule
    You are a Seeker who interacts with a Recommender to get a movie recommendation that suits your preferences well. You will be given a dialogue context, and you must follow the instructions below to interact with the Recommender:
    \\
    \\
    - The recommender may ask for your preference or recommend a movie to you. \\
    - In the beginning, express your general preference on movies based on your past reviews about movies you have watched before. \\
    - If you are recommended a movie which is not \{gt movie title\}, you should reject it with a reason based on your thought about the recommended movie. Also, express some common features of the movies you have watched before and you should be recommended (DO NOT explicitly mention the movie title!). \\
    - If you are recommended \{gt movie title\}, you should accept it as if you haven't watched it before, and end the conversation by generating [EOD] token. \\
    - Continue the conversation for at least six turns.
    \\ \\
    Here are your reviews about movies you have watched before: \\
    \{user persona\} \\ \\
    Some features of the movie you should be recommended: \\
    \{gt abstract\} \\ \\ 
    \{rec movie abstract\} \\ \\ 
    Here is the dialogue context: \\
    \{dialogue context\} \\
     \bottomrule
    \end{tabular}
    \caption{The prompt for Seeker simulator.}
    \label{tab:seeker_prompt}
\end{table*}




\begin{table*}
    \small
    \centering
    \begin{tabular}{p{14cm}}
    \toprule
    \textbf{Prompt} \\
    \midrule
    You are a Recommender who recommends a Seeker a movie that he/she will enjoy, among the three candidates and based on the dialogue context given. Follow the instructions below to complete the task:
    \\ \\
    - In the beginning of the conversation, engage with the Seeker to discover his/her movie preferences (regardless of the given three candidates). Follow this format: \\ 
    \, Think: (think about which question to ask the seeker) \\
    \, Recommender: (utterance that asks for the seeker's movie preference) \\
    - After some interactions, choose and suggest a movie from the three candidates and make the recommendation in the specified format: \\
    \, Think: (think about the seeker's movie preference based on the context) \\
    \, Movie: (Movie title (Year)) \\
    \, Recommender: (response to the seeker's utterance) \\
    - Do not recommend a movie that has been mentioned before in the dialogue context.
    \\ \\
    Here are the three movie candidates: \\
    \{k movies info\} \\ \\
    Here is the dialogue context: \\
    \{dialogue context\} \\ \\
    Think:
    \\

     \bottomrule
    \end{tabular}
    \caption{The prompt for Recommender simulator.}
    \label{tab:RG_prompt}
\end{table*}

{\renewcommand{\arraystretch}{1.35}
    \begin{table}[t] \begin{center}
    \small
    \setlength{\tabcolsep}{3pt}
    \begin{tabularx}{\linewidth}{X}
        \toprule
        \textbf{\textit{Dialogue context}} \\
        \textbf{Recommender}: Hey there \\
        \textbf{Seeker}: Hi! I'm looking for recommendations on a great holiday movie, do you have any recommendations? \\
        \textbf{Recommender}: For sure! First, do you have certain preferences or aversions? Like are you into musicals? \\
        \textbf{Seeker}: I do not really like musicals. \\
        \textbf{Recommender}: Good to know... I would have recommended White Christmas, it's a classic and a bit of a tradition in our family, but there are so many to choose from! The old Tim Allen Santa Clause movie is super fun and so is Home Alone. \\
        \textbf{Seeker}: Is the Tim Allen Santa Clause movie funny? I would love to watch a funny movie. \\
        \textbf{Recommender}: Yes it is! It's a heart warming family comedy :) \\
        \textbf{Seeker}: I accept this recommendation and would love to watch it with my family! Thank you for your help. \\
        \textbf{Recommender}: You're very welcome! I hope you enjoy it! \\
        \textbf{Seeker}: If we finish this movie early, do you have children's Christmas movie to recommend? So far our favorite is Elf. \\
        \midrule
        \textbf{\textit{Model responses}} \\
        \textbf{GPT-3.5}: I would recommend the movie "The Polar Express" as a great children's Christmas movie. It's a heartwarming and magical adventure that is perfect for the holiday season. Enjoy watching it with your family! \\
        \textbf{BART-ReDial}: Yes, I love Elf! It's a great family movie! \\
        \textbf{BART-\ourdata{} (Ours)}: How about watching `The Lion King'? It's a classic animated film with a heartwarming story and memorable characters. It's perfect for family viewing and has a great message about family and friendship. \\
        \midrule
        \textbf{Ground-truth}: I would absolutely recommend Rise of the Guardians it's a kid's movie, but definitely funny so adults can enjoy too. \\
        \bottomrule
    \end{tabularx}
    \vspace{-3pt}
    \caption{
        A Cherry-picked dialogue example from \ourdata{} dataset. This shows the difference in response generation between BART-ReDial and BART-\ourdata{}. This case illustrates that BART-\ourdata{} offers greater informativeness and explainability for the movie.
    }
    \vspace{-10pt}
    \label{tab:case_study_cherry}
\end{center}\end{table}}

{\renewcommand{\arraystretch}{1.35}
    \begin{table}[t] \begin{center}
    \small
    \setlength{\tabcolsep}{3pt}
    \begin{tabularx}{\linewidth}{X}
        \toprule
        \textbf{\textit{Dialogue context}} \\
        \textbf{Recommender}: Hello \\
        \textbf{Seeker}: hello  \\
        \textbf{Recommender}: What can I help you with  \\
        \textbf{Seeker}: Can you help recommend a movie trailer to watch?  \\
        \textbf{Recommender}: I would recommend horror movies, but I would like to know what movies you like first  \\
        \textbf{Seeker}: Oh, I love action movies.  \\
        \textbf{Recommender}: The types with very serious actors (Stallone, etc) or, like Guardians of the Galaxy?  \\
        \textbf{Seeker}: I like both, but I'm looking for something super hero related, like Guardians of the Galaxy.  \\
        \textbf{Recommender}: Oh! I have a few ideas of what you have in mind. Are comic book characters interesting at all?  \\
        \textbf{Seeker}: Yes! I love comic book characters, like Spider man, Ironman and Captain America.  \\
        \textbf{Recommender}: Okay. Would you like to watch this trailer for the Avengers?  \\
        \textbf{Seeker}: Oh that would be great!  \\
        \midrule
        \textbf{\textit{Model responses}} \\
        \textbf{GPT-3.5}: I would recommend you watch the trailer for "Avengers: Endgame." It features all your favorite comic book characters like Iron Man, Captain America, and Spider-Man, and it's a super hero-related movie just like Guardians of the Galaxy. Enjoy watching the trailer! \\
        \textbf{BART-ReDial}: Okay. I would also recommend `The Avengers (2012)' \\
        \textbf{BART-\ourdata{} (Ours)}: How about `The Avengers (2012)'? It's an action-packed superhero movie with a great cast and some really interesting characters. \\
        \midrule
        \textbf{Ground-truth}: I haven't actually watched it yet. It is the one with that Thanos character and the Justice league, right? \\
        \bottomrule
    \end{tabularx}
    \vspace{-3pt}
    \caption{
        Lemon-picked dialogue example from \ourdata{} dataset. This shows that BART-\ourdata{} recommends Avengers again which is recommended already in the previous dialog.
    }
    \vspace{-10pt}
    \label{tab:case_study_lemon}
\end{center}\end{table}}

\end{document}